# Object Recognition in Different Lighting Conditions at Various Angles by Deep Learning Method

Imran Khan Mirani[a], Chen Tianhua[b], Malak Abid Ali Khan[c], Syed Muhammad Aamir[c], Waseef Menhaj[d]

[a]School of Information and Computer Engineering, Beijing University of Technology, Beijing, 100124, China
[b]School of Artificial Intelligence, Beijing Technology & Business University, Beijing, 100048, China
[c]School of Automation, Beijing Institute of Technology, Beijing, 100081, China
[d]Department of Electronic & Information Engineering, Anhui Normal University, Wuhu, 241000, China

**Abstract:** Existing computer vision and object detection methods strongly rely on neural networks and deep learning. This active research area is used for applications such as autonomous driving, aerial photography, protection, and monitoring. Futuristic object detection methods rely on rectangular, boundary boxes drawn over an object to accurately locate its location. The modern object recognition algorithms, however, are vulnerable to multiple factors, such as illumination, occlusion, viewing angle, or camera rotation as well as cost. Therefore, deep learning-based object recognition will significantly increase the recognition speed and compatible external interference. In this study, we use convolutional neural networks (CNN) to recognize items, the neural networks have the advantages of end-to-end, sparse relation, and sharing weights. This article aims to classify the name of the various object based on the position of an object's detected box. Instead, under different distances, we can get recognition results with different confidence. Through this study, we find that this model's accuracy through recognition is mainly influenced by the proportion of objects and the number of samples. When we have a small proportion of an object on camera, then we get higher recognition accuracy; if we have a much small number of samples, we can get greater accuracy in recognition. The epidemic has a great impact on the world economy where designing a cheaper object recognition system is the need of time. First, gather enough samples (custom) as our dataset and use the appropriate Yolov2 model to complete the training and testing. We rendered two separate groups of objects and obtained images of each object as a training set, each of which can be cropped and rotated using different angles of the image with features of an object, for training and verification. The use of neural networks to identify objects increases recognition rates. We present consequences on the custom dataset display the value of the borders of flexible objects, particularly with rotated and non-rectangular objects.

## I. INTRODUCTION

An individual can recognize just about any kind of object around him. Even so, machine recognizing objects accurately is a too lengthy operation. Suppose a mature adult can easily differentiate between the objects, but it's difficult for children to discern the difference between them. We teach them examples and they'll begin to learn about them. The same principle applies to computers with objects and the expected results will be learned and given. Camera-based object recognition is one of the fastest developing research fields of computer vision and machine learning. Object detection has been researched and implemented at different locations on image and video. These include computer vision, robotics, automation, design, and farming. It is by looking and seeing that we come to know the world in which we live. The natural world is lined up with countless forms of objects and perceptions. In other words, perception is a way of gaining knowledge for the world around us. It remains a mystery to understand exactly how the visual system functions, even though physiologists have been studying the phenomenon for decades. When we talk about vision, we have the specific and abstract form of computer vision by replacing the living creature with a digital instrument. It can be defined as the process of computers analyzing digital images or videos and gaining an understanding of them at a high level [1].

Deep learning is an apprenticeship technique. It's a high-dimensional data reduction for both the input and output models. We arrange deep learning as artificial intelligence→machine learning→deep learning. The deep learning extracts feature shapes its architecture all by itself into multiple layers. There is a bundle of applications within artificial intelligence in deep learning. Few are noted here: Automation of robotics and image processing in the recognition of live objects. Deep learning has had a great impression in recent research work and it is one of the sections on computer vision's hot and breathtaking research subject [2]. It is strongly thinking and an inner class of artificial intelligence, while machine learning can generate a function of multiple layers of neural networks from given input data. It is the state-of-the-art architecture of artificial intelligence, which is one of the main research areas at the moment. The researchers have been extremely interested in solving highly challenging tasks through deep learning models and algorithms such as the implementation of artificial intelligence in "real-time object recognition" imagery related to their classification and detection at the very big data level. Object recognition on the surface is a fairly difficult task from now on we want to identify the architecture of deep learning that will enhance accurately the detection of a particular object.

We often address old traditional and modern approaches that are unacceptable as the new methods of deep learning, such as hand-crafted methods and methods of color detection. Deep learning through the surface camera at the moment depends on new, different techniques. We can use some of the new methods that are quicker and more efficient according to the requirements of our day. Just like faster-RCNN and YOLO (real-time detection systems), there are plenty of methods available. We may also find perennial approaches where the researchers don't need deep learning neural networks. Instead, they recommend using texture and color detection to identify objects. The old methods aren't compatible with enormous data set knowledge. Recognition of real-time objects has a major demand and a very demanding challenge. I have decided to focus on object recognition after all the research and brainstorming. I expect to incorporate a deep learning architecture and one of the best methods of recognizing various surface objects. The convolution neural network and Yolo model is actual, and the most popular and well-suited architecture for object recognition.

## II.    RELATED WORK

Object detection methods adopt the feature-extraction-plus-classifier approach before 2012. First, to describe the object correctly, people must consciously identify unique features for a category of objects. Upon extracting sufficient features from the training dataset, a vector of features will represent the object that is used to train a classifier in training time it can also be used to fulfill the role of detection in test time. If people choose to create a detector of multiple objects, further attention is required in choosing a general feature so that the specific feature can match different objects. The downside is evident because the description of the function is so complex and when the new object is added to the detection list, the concept becomes difficult to extend. While the accuracy of detection is unsatisfactory.

In the 2012 Large Scale Visual Recognition Challenge (ILSVRC, 2012), the CNN-based model from Krizhevsky outperformed all of the other models by a major margin [3]. Although the identical CNN framework was introduced in the 1990s, the lack of training examples and poor hardware had obscured its ability. A subset of ImageNet [4] has been used for classification in ILSVRC (2012) which contains 1.2 million images from 1,000 categories. In turn, the GPUs became more powerful. Krizhevsky's experiment proved CNN's powerful capability in image recognition with ample training images and strong GPUs. The key benefit CNN has over conventional approaches is the opportunity during the training process to create function filters. Researchers are set free from the laborious development of software. CNN, therefore, has ahead of description, and ahead of regression. Two heads work together at test time, one for estimating class score and one for location. From 2012 to 2015, work subsequently drafted regression heads for the latest remarkable CNNs, such as Overfeat-Net [5], VGG-Net [6], and ResNet [7], reducing the single object position error from 34% to 9%. CNN-based methods worked very well in the role of localization-plus classification. Researchers later started moving on to multiple object detection activities in 2014. A single image contains more than one object, and the machine needs to figure out the class and location of each object. Several proposals-plus-classification solutions focused on area CNN appeared [8-11]. The key concept is to use regional proposal methods to produce applicant regions that can include artifacts, and then identify each region later. In practice, these strategies to region-proposals-plus-classification may achieve very high accuracy [5-7]. The area proposal portion, however, is very time-consuming, which slows down the speed of the entire system. The sluggish detection speed prohibits the application of these methods to time-critical applications such as auto driving, surveillance system, etc. A unified model of object detection, YOLO [12], was recently proposed by Joseph. It frames detection as a question of regression. The model regresses directly from the input image to a tensor representing the score of object class and the digits of the position of each object. The input images only need to go over the network once and the model processes images faster as a result. YOLO has achieved an accuracy of more than 50 percent in real-time detection on data set VOC 2007 and 2012, making it the best option for any real-time object detection tasks. Tensorflow plays a crucial role in the creation and implementation of CNNs, which has incorporated most specific units into the deep learning system (Shi, Wang, Xu, & Chu, 2016), and is also independent of the C++, python-based open-source platform for DNNs (Flores, Barrón-Cedeño, Rosso, & Moreno, 2011), and CUDA (Kirk, 2007) model. One important feature is flexible usability which makes it easy to install the very same codes without alteration to numerous CPUs and GPUs. Another reason to use Tensorflow as our platform is that the library Tensorflow is the DNN's description of the software system that is commonly used. A deep learning system as the core function of the DNN models must be explicit before the DNNs model is implemented. It consists of two phases "Define" and "Run," namely, Define-and-Run (Tokui, Oono, Hido, & Clayton, 2015) in a traditional DNNs structure model such as Tensorflow.

## III.    OVERVIEW OF PROPOSED METHODOLOGY

Figure provides an outline of our proposed Real-time object recognition by deep learning. Generally can be divided into the following important phases:

1.  Data collection
2.  Training Yolov2 for object detection.
3.  Data Annotation & labeling
4.  Augmentation of data.
5.  Testing & Evaluation
6.  Outcome

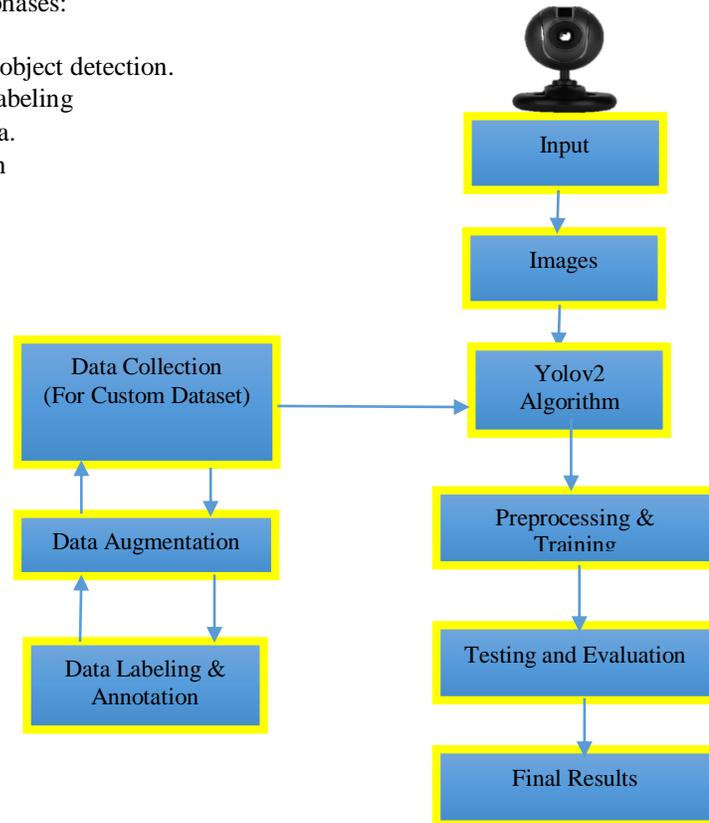

Figure 1. Overview of Proposed Methodology

The proposed architecture is primarily developed by deep learning for the recognition of real-time objects. Object recognition is a key technology behind driverless cars, facial recognition, and counting crowds. This approach additionally takes full advantage of existing computing resources and can be applied locally on low computing devices. This approach is adaptable to different variations in the atmosphere, i.e. variation in lighting, deformation, background noise, variation in perspective, occlusion (in whole or in part), motion blur, changes in size, and so on. In dim light, the system works better. The data collection we are collecting comprises two groups. Table 3.1 reveals we have two study participants. Since we collected the data by ourselves in this study, there are two classes that we collect, each class represents one group.

Table 1. Classes in our Experiment

| | |
|---|---|
| 1 | Cell |
| 2 | Laptop |

Should pay attention to the following points:

*   The picture is clear and the motion is small, thereby preventing the dynamic blurring of the background.
*   Different locations of an object in a picture.

- Objects are large enough in scale and require photographs from various distances to be taken.
- Objects are big enough in shapes, indicating that they are taken under various words and actions.

Data labeling is a job that demands a lot of manual work. Used a LabelImge tool to mark Rectangular shapes based on objects where the labeling procedure is shown in Figures 2(a) and 2(b).

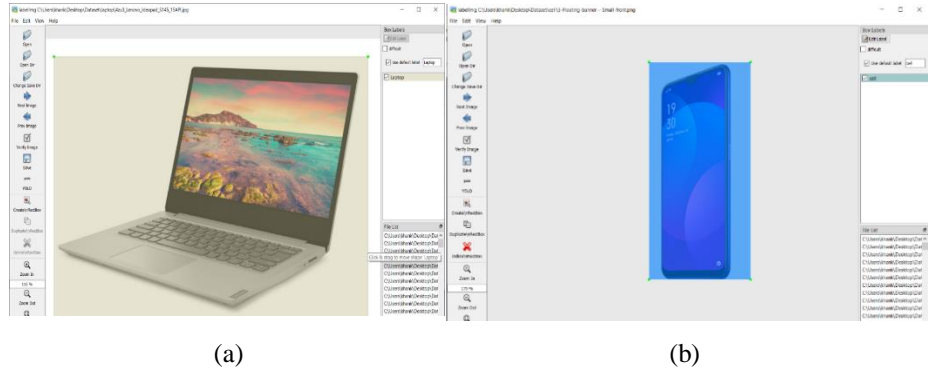

(a)                                        (b)

Figure 2. (a) Labelling the laptop (b) Labelling the cellphone

Got initial images that were marked after naming the data collection. That number is not enough for testing, however, if the number of our data set is very small, then the model testing will be insufficient, which will impact the model's final accuracy. We tried a method to boost the data and increase the amount of data to render the training dataset fully qualified.

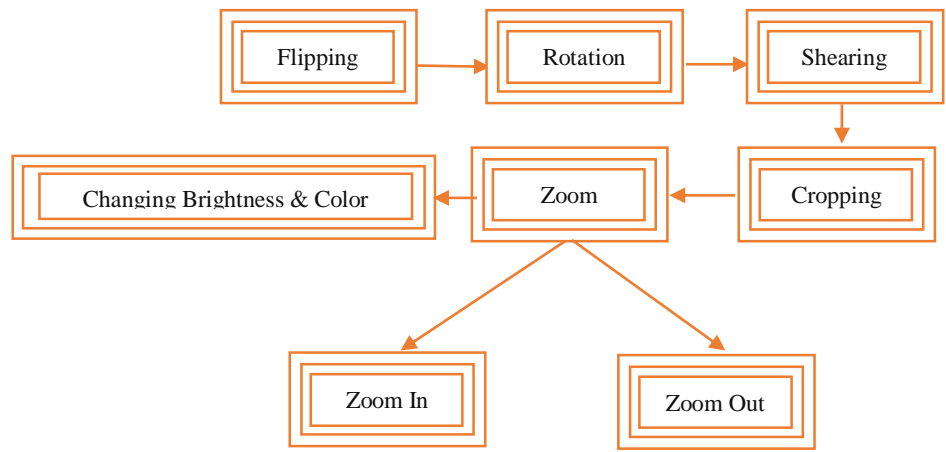

Figure 3. Flowchart of data augmentation

Use the rectangular shape to mark the object so that when the item is tilted the object can be recognized. Since the photographs are randomly chosen frames from the web and taken from the camera, the untitled object occupies the main part of the dataset. For instance, if one of the 100 images contains an image, the object is labeled. While the model does not consider the labeled object, the precision will exceed 99%. That won't predict the artifacts being tilted very well. So, we have to increase the number of objects tilted. We will then perform random cropping of the image. The aim of this is to allow an object to occupy various positions in the image to increase the object's size and location. The location of the object and its ratio to the image have changed after cropping. We need to scale the cropped image down to our desired size after cropping so that it can be used as training data. We convert the RGB color to the HSV color space and change the saturation (S) and value (V) part randomly, considering that different cameras would have a different color when shooting to simulate the color difference.

## IV.    TRAINING & DETECTION WITH YOLOV2

To apply the efficient, fast, and lightweight Yolov2 in the proposed Real-time object Recognition by deep learning method. The following steps are followed for the training object detection model. First, we need a suitable object detection dataset to train a model. Then prepare the dataset for annotation which includes coordinates for bounding boxes and labels for image classification. Training Parameter Selection: This part is a tedious task too. One should make many attempts to find the appropriate training parameters to find the best parameters. Some of the training parameters include image quality, batch size, learning rate, number of epochs, and decay in learning rates. Besides, training should be effective after proper selection of training parameters and training is started so that precision increases over time/increase of epochs and loss value decrease. If each preliminary step is set up correctly, it will achieve equilibrium at one point, i.e. the accuracy and loss will stabilize. That is a strong indicator of good training and it means that the trained object detection model is ready for use in the detection of target objects.

Ultimately, once the model is successfully trained we can use our trained model to detect items of interest (target items). The incoming images needed to resize pixels depending on the shape of the data that you set during the yolov2 training. The picture must be resized and these are the only and most important steps in preprocessing. The resized images will then serve as inputs for qualified model yolov2. For detection, the model takes certain input images and returns a set of object detection results for the input image after processing. The effects of visual detection include the target information and the objects selected in the input image. The finally resulting information includes objects classifications (Labels + Predictions score) and bounding box coordinates. Our methodology's detection module can recognize the object by giving results in label type, prediction score, and bounding box. Object detection on each frame is computationally costly, and the trained model takes time to process each input image. Instead of selecting interesting parts of an image, regression-based algorithms run through the entire image and estimate classes and bounding boxes in one run. Since the name of YOLO means exactly that the algorithm looks through the entire image just once and places the bounding boxes on each object in the image that the algorithm thinks there is a possibility that this object exists in that region of the image and then removes the bounding boxes with the lower probability percentage of presence, the threshold the vary from app to app. The default threshold value of YOLOv2 eliminates any bounding box predictions below 60 percent and only the highest bounding box confidence percentages remain.

## V.    ENVIRONMENT

The experiment's hardware setting is illustrated in Table. 3.3, Ubuntu 16.04 was the operating system and the templates were introduced in the Darknet application environment.

Table 2. The hardware environment. GPU graphics processing unit, CPU central processing unit.

| Hardware | Environment |
|---|---|
| Computer | ThinkPad Laptop |
| CPU | Intel(R) Core (TM) i7-5500U CPU @ 2.40GHz 2.39 GHz |
| ++`GPU | NVIDIA GeForce 940M |

## VI.    IMPLEMENTATIONS AND RESULTS

Our proposed solution is based on deep learning YOLO framework for Real-time object detection. As we needed high accuracy and high-speed detection we tried to use YOLOv2 [13]. Our proposed method of computer software is developed to check the validity and opportunity of our approach to low computational devices. A custom dataset was developed in the Yolo data set format for deep learning-based object detection. The model is trained on a Laptop with GeForce 940 M GPU using our custom data collection. The tests of the model for object detection are also being tested on Windows and Ubuntu. To assess our proposed methodology and verify its findings, we evaluated our application on target and qualified artifacts. The following section discusses the development of the application and description of the experiment and the results of the experiment.

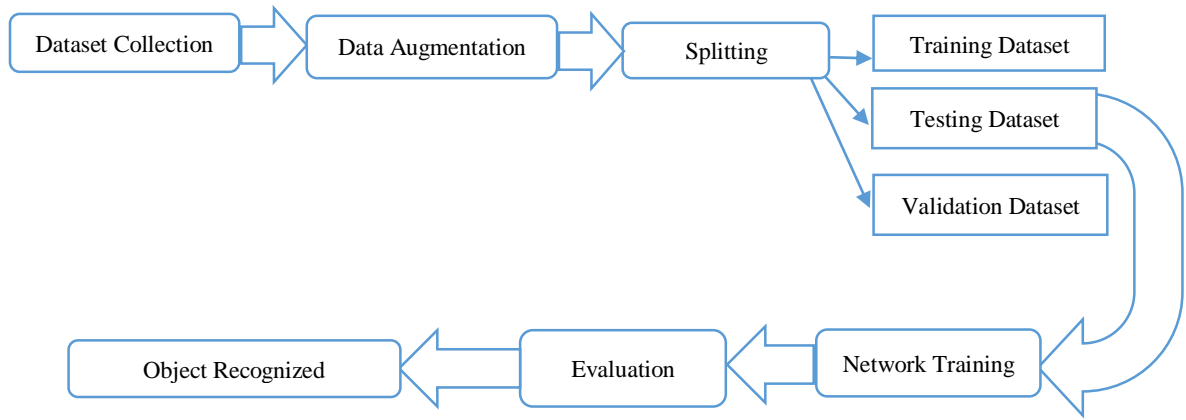

Figure 4. Process of Real-time Custom object Recognition

## Custom Object Detection Data Preparation

Collected 1000 images to create our object detection dataset and train the model. For our dataset, select the format Yolo. Our dataset has two types, i.e. Cell phone and Laptop. For every class, we captured 500 pictures. To assist the extractor of the feature in extracting and understanding the critical feature from the points of interest, we shot the images from various angles/directions, multiple sizes, and under specific lighting conditions. The main goal is to allow our successfully trained object detection model to recognize and detect these selected objects in input images at the various possible sizes, from different directions, and finally in any possible variation in illumination.

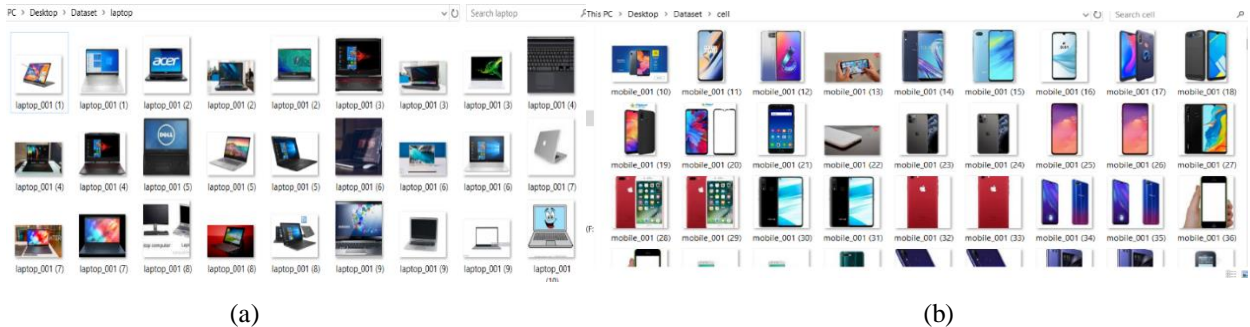

(a)                                                                              (b)

Figure 5. (a) Dataset collection of laptops (b) Dataset collection of cellphones

## Preconditioning and Training

Capturing pictures by itself is not enough to train a model for object detection. In our custom object detection dataset, annotations are required specifying the bounding box coordinates and classifications for each selected object. Our object detection dataset was organized in Yolo format. First, all you can use image size is up to you otherwise Yolo will keep aspect ratio, i.e. image 1280x720 will be resized to 416x234, then inserted into the 416x416 network and then annotated all images by labeling and creating bounding box coordinates with LabelImage graphical annotation toolset. In the Yolo data set design, the images and their corresponding annotations are ordered. Annotations that include coordinates and object labels are stored in the TXT file format.

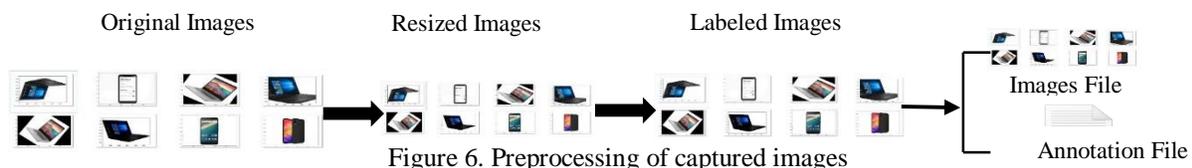

Figure 6. Preprocessing of captured images

Picked images for training from the pre-processed files and selected annotated images for validation. To train the detector of objects we used test set and validation sets from our custom dataset. Ultimately, the test set was used to calculate the average accuracy of individual objects and mean average accuracy to check the overall accuracy and performance of our trained object detection model. To get an effective and accurate detection model, we carefully set all training parameters to train our deep learning-based detector i.e. Yolov2. During the training of the model, specifications are set following.

Table 3. Parameters selection.

| Yolov2 cfg-file: | subdivisions= 16 | steps=4800,5400 | Network size: width=416 height=416 | classes=2 |
|---|---|---|---|---|
| filters= (2 + 5) x3 | channels=3 | momentum=0.9 | decay=0.0005 | angle=0 |
| saturation = 1.5 | exposure = 1.5 | hue=0.1 | learning rate=0.001 | batch=64 |

YOLO algorithm is among the frameworks used for real-time object detection and tracking in most apps. The problem of object detection is described as a problem of regression compared to the problem of classification, which uses different solutions like RCNN [14], Fast-RCNN [15], Faster-RCNN [16], and Mask RCNN [17]. These networks are therefore slower than YOLO when it comes to object detection as RCNN uses a pipeline in multiple steps to perform this function. This can be slow to run and also difficult to automate because each part has to be trained independently. Started the training with the above parameters and managed to obtain an effective model for object detection.

Table 4. Information of Dataset

| Laptop | Cell Phone | |
|---|---|---|
| 0.81 (81%) | 0.79 (79%) | Average Precision (AV) |
| 0.80 (80%) | | Mean Average Precision (mAP) |

## Training Results and Validation

AP and mAP values are the best precision measures. The higher the mAP value indicates the model is reliable and reliably strong enough and can be generalized. Table 5 displays the AP and mAP values obtained for the specified custom objects throughout Yolov2 training. The Yolov2 showed a steady rise in accuracy and a decline in loss during training. Convergence is detected after it has passed roughly epochs. Both the accuracy and the loss values (mAP) have become stable. The alignment and consistent principles for effective training are strong indications.

Table 5. Displays values on Training Data

| Name | Explanation | Usage |
|---|---|---|
| Learning rate | Learning rate decides how fast the error goes to the global minimum error | Training |
| Momentum | Momentum speeds up the learning rate | -do- |
| Decay | Decay is a parameter to avoid overfitting Training | -do- |
| Batch size | How many images are in one batch? In training, there is always a batch of images that go through the network together and use average error to update the network | -do- |
| Total batch | How many batches of images are to be trained? | -do- |

| Class Threshold | To eliminate detections whose class scores are less than threshold | -do- |
|---|---|---|

The graph of the loss curve appears in Figure 7. The resulting graph clearly shows the gradual decrease in loss value and finally became stable, which is a direct indicator of our object detection model's effective training.

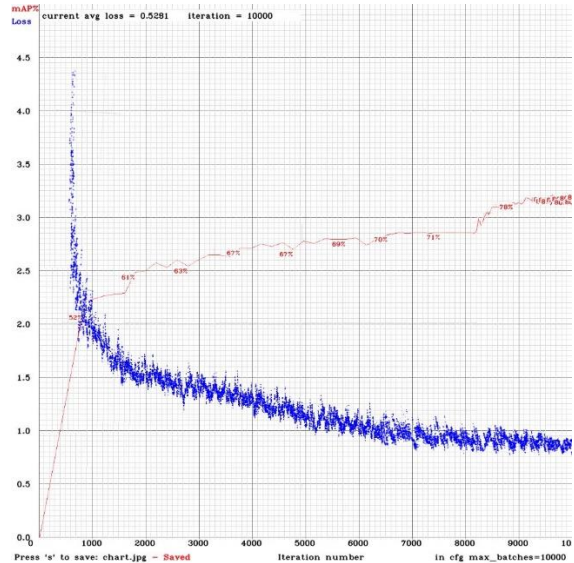

Figure 7. YOLOv2 Training

## Detection Results

The object detection model correctly identified and acknowledged the target object with a high score prediction. Below are some figures which clearly show our trained model's detection performance.

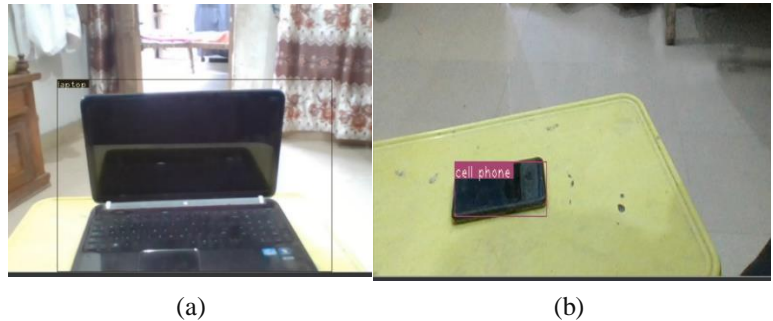

(a)                              (b)

Figure 8. (a) Real-Time laptop Detection (b) Real-Time cellphone Detection

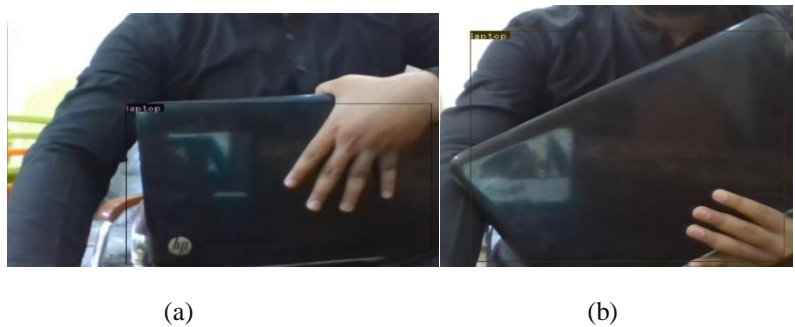

(a)                              (b)

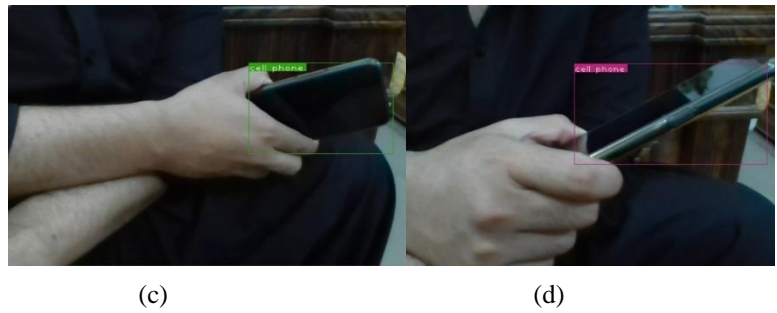

Figure 9. (a) Straight position of the laptop (b) Rotated position of the laptop (c) Straight position of the cellphone (d) Rotated position of the cellphone

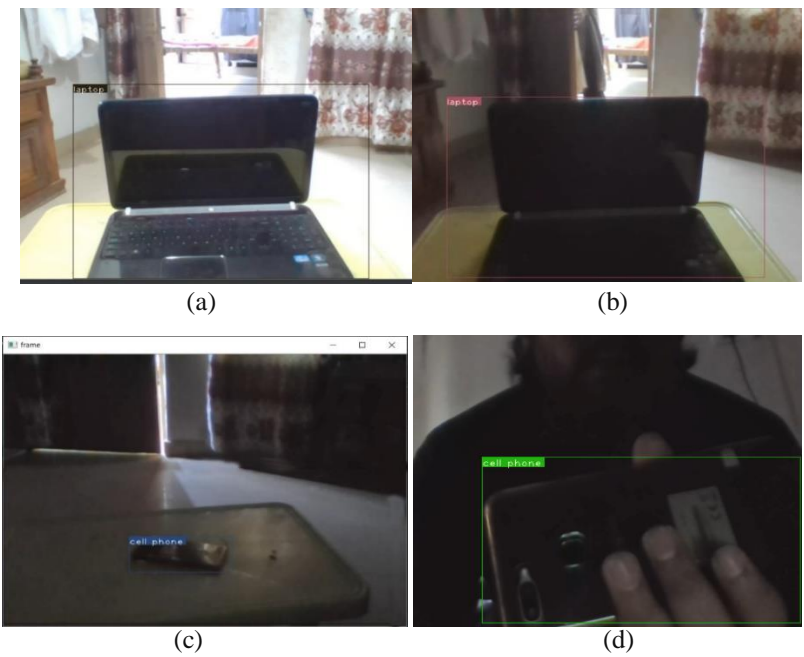

Figure 10. (a) Laptop detection in bright light (b) Laptop detection in dam light (c) Cellphone detection in dam light (d) Cellphone detection in no light

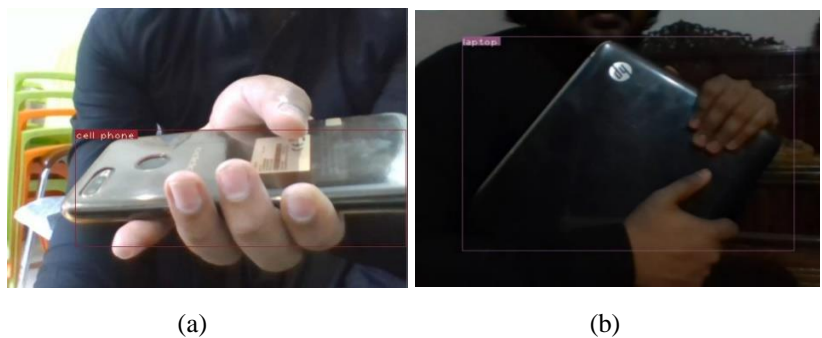

Figure 12. (a) Occlusion handling in bright light (cellphone) (b) Occlusion handling in no light (laptop)

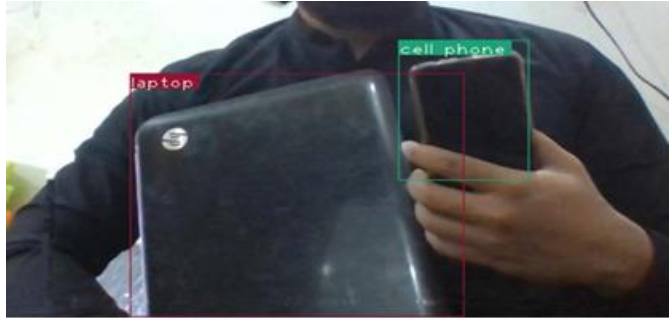

Figure 13. Multi Real-time object recognition in occluded environment

## VII.    CONCLUSION AND FUTURE WORK

The purpose of this study is to define each object's identity before a camera. The main research is the description of an object and the effect on the accuracy of the object's proportion. We're proposing two classes; each class describes classification. We collaborated with Custom data sets on one model under the same concept structure and summarized the best one as our final one. We find that our model can detect and recognize objects very well, using deep learning. After the object recognition has been completed our key contributions are summarized as we've found that an object's proportion influences accuracy. While an object is closer to the camera, the object is bigger, the software confidence tends to higher, and the result of recognition is good; when the object is far from the camera, the confidence slowly reduces; the device cannot detect it until the object is too far away. And we notice that our model has some limitations for object detection. Our model performs better than others according to obtained results which has greater usability in surveillance. Our model has a higher accuracy of 80% for different object recognition projects, and each class is well recognized.

In the future, our project will be expanding considerably, the results of our model are well-validated, and this does not extend to any form of training data set. We will refine our model in the future based on the current results of this project so that the model can be applied to various data sets. We may use the same training dataset to check more models like Yolov3 for instance. These can also include models on various systems, not limited to Tensorflow, to help pick a more robust model for object recognition. We should add more principles that are not only limited to object recognition but also include gender identity and facial expressions. That needs us to innovate our datasets and models, of course. We're trying to use other neural networks to attain object recognition, not only using convolution neural networks. We may replace the yolov2 with yolov5 or any other well-developed model according to future needs.

**Acknowledgments**

Special thanks to the Editor, Associate Editor, and anonymous reviewers for improving the quality of this research work.

**Funding Statement**

This work is supported by the National Natural Science Foundation of China (61702020), Beijing Natural Science Foundation (4172013), and Beijing Natural Science Foundation Haidian Primitive Innovation Joint Fund (L182007) in 2019.

**Conflicts of Interest**

The authors declare that they have no conflicts of interest to report regarding the present study.